\documentclass[11pt,a4paper]{article}
\usepackage[hyperref]{emnlp-ijcnlp-2019}
\usepackage{standalone}
\usepackage{mystuff-nlp}
\usepackage{times}
\usepackage{tikz}
\usepackage{url}
\usepackage{latexsym}
\usepackage{color}
\usepackage{float}
\usepackage{amsfonts,amssymb,amsmath,bm,bbm}
\usepackage{algpseudocode}
\usepackage{graphicx,subfigure}
\usepackage{booktabs}
\usepackage{multirow}
\usepackage{verbatim}

\usepackage{enumitem,array}
\usepackage[titlenumbered,ruled]{algorithm2e}

\usepackage{setspace}

\newcommand{\example}[1]{`#1'}
\newcommand{\sysname}{\textsc{Av-Kmeans}}

\aclfinalcopy 

\title{Dialog Intent Induction with Deep Multi-View Clustering}

\author{Hugh Perkins \and Yi Yang\\
	ASAPP Inc.\\
	New York, NY 10007\\
	{\tt \{hp+yyang\}@asapp.com}
        }

\date{}

\begin{document}
\maketitle

\fbox{%
  \parbox{0.4\textwidth}{
  \small
    Original version: Proceedings of EMNLP 2020. We have added an appendix
    which includes experiments on a slightly larger askubuntu dataset, and incorporating
    several post-publication code bug-fixes. In addition, the number of questions in the
    AskUbuntu dataset has bee corrected, along with the associated cluster sizes.
    While the experimental results are largely similar, we
    update the paper for the sake of completeness.
  }%
}
\vspace{3em}

\begin{abstract}
We introduce the dialog intent induction task and present a novel deep multi-view clustering approach to tackle the problem. Dialog intent induction aims at discovering user intents from user query utterances in human-human conversations such as dialogs between customer support agents and customers.\footnote{We focus on inducing abstract intents like \texttt{BookFlight} and ignore detailed arguments such as \emph{departure date} and \emph{destination}.}
Motivated by the intuition that a dialog intent is not only expressed in the user query utterance but also captured in the rest of the dialog, we split a conversation into two independent views and exploit multi-view clustering techniques for inducing the dialog intent. In particular, we propose alternating-view k-means (\sysname) for joint multi-view representation learning and clustering analysis. The key innovation is that the instance-view representations are updated iteratively by predicting the cluster assignment obtained from the alternative view, so that the multi-view representations of the instances lead to similar cluster assignments.
Experiments on two public datasets show that \sysname~can induce better dialog intent clusters than state-of-the-art unsupervised representation learning methods and standard multi-view clustering approaches.\footnote{The data and code are published at \url{https://github.com/asappresearch/dialog-intent-induction}.}


\end{abstract}

\section{Introduction}
\label{sec:intro}

Goal-oriented dialog systems assist users to accomplish well-defined tasks with clear intents within a limited number of dialog turns. They have been adopted in a wide range of applications, including booking flights and restaurants~\cite{hemphill1990atis, williams2012belief},  providing tourist information~\cite{kim2016fifth}, aiding in the customer support domain, and powering intelligent virtual assistants such as Apple Siri, Amazon Alexa, or Google Assistant. The first step towards building such systems is to determine the target tasks and construct corresponding ontologies to define the constrained set of dialog states and actions~\cite{henderson2014word, mrkvsic2015multi}.

\begin{figure}
\small
\centering
\noindent\fbox{%
    \parbox{6.8cm}{%
        \textbf{Customer 1}: \textit{A wireless charging case is fancy and all but can we get a ``find my airpod'' feature going?} \\
        \textbf{Agent 1}:  \textit{If you have lost your AirPods, Find My iPhone can help you locate them.}
    }%
}

\noindent\fbox{%
    \parbox{6.8cm}{%
        \textbf{Customer 2}: \textit{hey man I lost and miss my airpods plz help me!} \\
        \textbf{Agent 2}:  \textit{Hi there! With iOS 10.3 or later, Find My iPhone can help you locate missing AirPods.}
    }%
}
\caption{Two dialogs with the \texttt{FindAirPods} user intent. The user query utterances of the two dialogs are lexically and syntactically dissimilar, while the rests of the dialogs are similar.}
\label{fig:dialog-example}
\end{figure}

Existing work assumes the target tasks are given and excludes dialog intent discovery from the dialog system design pipeline. Because of this, most of the works focus on few simple dialog intents and fail to explore the realistic complexity of user intent space~\cite{williams2013dialog, budzianowski2018multiwoz}. The assumption puts a great limitation on adapting goal-oriented dialog systems to important but complex domains like customer support and healthcare where having a complete view of user intents is impossible. For example, as shown in~\autoref{fig:dialog-example}, it is non-trivial to predict user intents for troubleshooting a newly released product in advance. To address this problem, we propose to employ data-driven approaches to automatically discover user intents in dialogs from human-human conversations. Follow-up analysis can then be performed to identify the most valuable dialog intents and design dialog systems to automate the conversations accordingly.

Similar to previous work on user question/query intent induction~\cite{sadikov2010clustering, haponchyk2018supervised}, we can induce dialog intents by clustering user query utterances\footnote{We treat the initial user utterances of the dialogs as user query utterances.} in human-human conversations.  The key is to learn discriminative query utterance representations in the user intent semantic space. Unsupervised learning of such representations is challenging due to the semantic shift across different domains~\cite{nida2015componential}. We propose to overcome this difficulty by leveraging the rest of a conversation in addition to the user query utterance as a weak supervision signal. Consider the two dialogs presented in~\autoref{fig:dialog-example} where both of the users are looking for how to find their AirPods. Although the user query utterances vary in the choice of lexical items and syntactic structures, the human agents follow the same workflow to assist the users, resulting in similar conversation structures.\footnote{Note this is not always the case. For the same dialog intent, the agent treatments may differ depending on the user profiles. The user may also change intent in the middle of a conversation. Thus, the supervision is often very noisy.}

We present a deep multi-view clustering approach, alternating-view k-means (\sysname), to leverage the weak supervision for the semantic clustering problem. In this respect, we partition a dialog into two independent views: the user query utterance and the rest of the conversation. \sysname~uses different neural encoders to embed the inputs corresponding to the two views and to encourage the representations learned by the encoders to yield similar cluster assignments. Specifically, we alternatingly perform k-means-style updates to compute the cluster assignment on one view and then train the encoder of the other view by predicting the assignment using a metric learning algorithm~\cite{snell2017prototypical}. Our method diverges from previous work on multi-view clustering~\cite{bickel2004multi,chaudhuri2009multi,kumar2011co}, as it is able to learn robust representations via neural networks that are in clustering-analysis-friendly geometric spaces. Experimental results on a dialog intent induction dataset and a question intent clustering dataset show that \sysname~significantly outperforms multi-view clustering algorithms without joint representation learning by $6$--$20$\% absolute F1 scores. It also gives rise to better F1 scores than quick thoughts~\cite{logeswaran2018efficient}, a state-of-the-art unsupervised representation learning method.

Our contributions are summarized as follows:
\begin{itemize}
\item We introduce the dialog intent induction task and present a multi-view clustering formulation to solve the problem.
\item We propose a novel deep multi-view clustering approach that jointly learns cluster-discriminative representations and cluster assignments.
\item We derive and annotate a dialog intent induction dataset obtained from a public Twitter corpus and process a duplicate question detection dataset into a question intent clustering dataset. 
\item The presented algorithm, \sysname, significantly outperforms previous state-of-the-art multi-view clustering algorithms as well as two unsupervised representation learning methods on the two datasets.
\end{itemize}

\section{Deep Multi-View Clustering}
\label{sec:model}

In this section, we present a novel method for joint multi-view representation learning and clustering analysis. We consider the case of two independent views, in which the first view corresponds to the user query utterance (query view) and the second one corresponds to the rest of the conversation (content view). 

Formally, given a set of $n$ instances $\{ x_i \}$, we assume that each data point $x_i$ can be naturally partitioned into two independent views  $x_i^{(1)}$ and $x_i^{(2)}$.  We further use two neural network encoders $f_{\phi_1}$ and $f_{\phi_2}$ to transform the two views into vector representations $\mat{x}_i^{(1)}, \mat{x}_i^{(2)} \in \reals^D$. We are interested in grouping the data points into $K$ clusters using the multi-view feature representations. In particular, the neural encoders corresponding to the two views are jointly optimized so that they would commit to similar cluster assignments for the same instances.

In this work, we implement the query-view encoder $f_{\phi_1}$ with a bi-directional LSTM (BiLSTM) network~\cite{hochreiter1997long} and the content-view encoder $f_{\phi_2}$ with a hierarchical BiLSTM model that consists of a utterance-level BiLSTM encoder and a content-level BiLSTM encoder. The concatenations of the hidden representations from the last time steps are adopted as the query or content embeddings.



\subsection{Alternating-view k-means clustering}

In this work, we propose alternating-view k-means (\sysname) clustering, a novel method for deep multi-view clustering that iteratively updates neural encoders corresponding to the two views by encouraging them to yield similar cluster assignments for the same instances. In each semi-iteration, we perform k-means-style updates to compute a cluster assignment and centroids on feature representations corresponding to one view, and then project the cluster assignment to the other view where the assignment is used to train the view encoder in a supervised learning fashion.

\begin{algorithm}[h]
  \small
  \setstretch{1.1}
  \SetKwInOut{Input}{Input}
  \SetKwInOut{Output}{Output} 
  \SetKwInOut{Parameter}{Parameter}
  \Input{two-view inputs $\{ (x_i^{(1)}, x_i^{(2)}) \}$; numbers of iterations $T$, $M$; \qquad number of clusters $K$}
  \Output{final cluster assignment $\{ z_i^{(1)} \}$}
  \Parameter{encoders $f_{\phi_1}$ and $f_{\phi_2}$}\
  Initialize $f_{\phi_1}$ and $f_{\phi_2}$ (\autoref{sec:model:init})\\
  $\{ z_i^{(1)} \}  \leftarrow \textproc{K-means}(\{ f_{\phi_1}(x_i^{(1)}) \}, K) $ \\
  \For{$t = 1, \cdots, T$}{
      // \texttt{project cluster assignment from view 1 to view 2}\\
      Update $f_{\phi_2}$ with pseudo training instances $\{(x_i^{(2)},  z_i^{(1)})\}$ (\autoref{sec:model:train}) \\
      Encode view-2 inputs: $\{ \mat{x}_i^{(2)} \leftarrow f_{\phi_2}(x_i^{(2)}) \}$ \\
      $\{ z_i^{(2)} \}  \leftarrow \textproc{K-means}(\{ \mat{x}_i^{(2)} \}, K, M, \{ z_i^{(1)} \} ) $ \\
       \quad \\
      // \texttt{project cluster assignment from view 2 to view 1}\\
      Update $f_{\phi_1}$ with pseudo training instances $\{(x_i^{(1)},  z_i^{(2)})\}$ (\autoref{sec:model:train}) \\
      Encode view-1 inputs: $\{ \mat{x}_i^{(1)} \leftarrow f_{\phi_1}(x_i^{(1)}) \}$ \\
      $\{ z_i^{(1)} \}  \leftarrow \textproc{K-means}(\{ \mat{x}_i^{(1)} \}, K, M, \{ z_i^{(2)} \} ) $ \\
  }
 \caption{alternating-view k-means}
 \label{algo:av-kmeans}
\end{algorithm}

\begin{figure}[t]
\centering
\includegraphics[scale=.4]{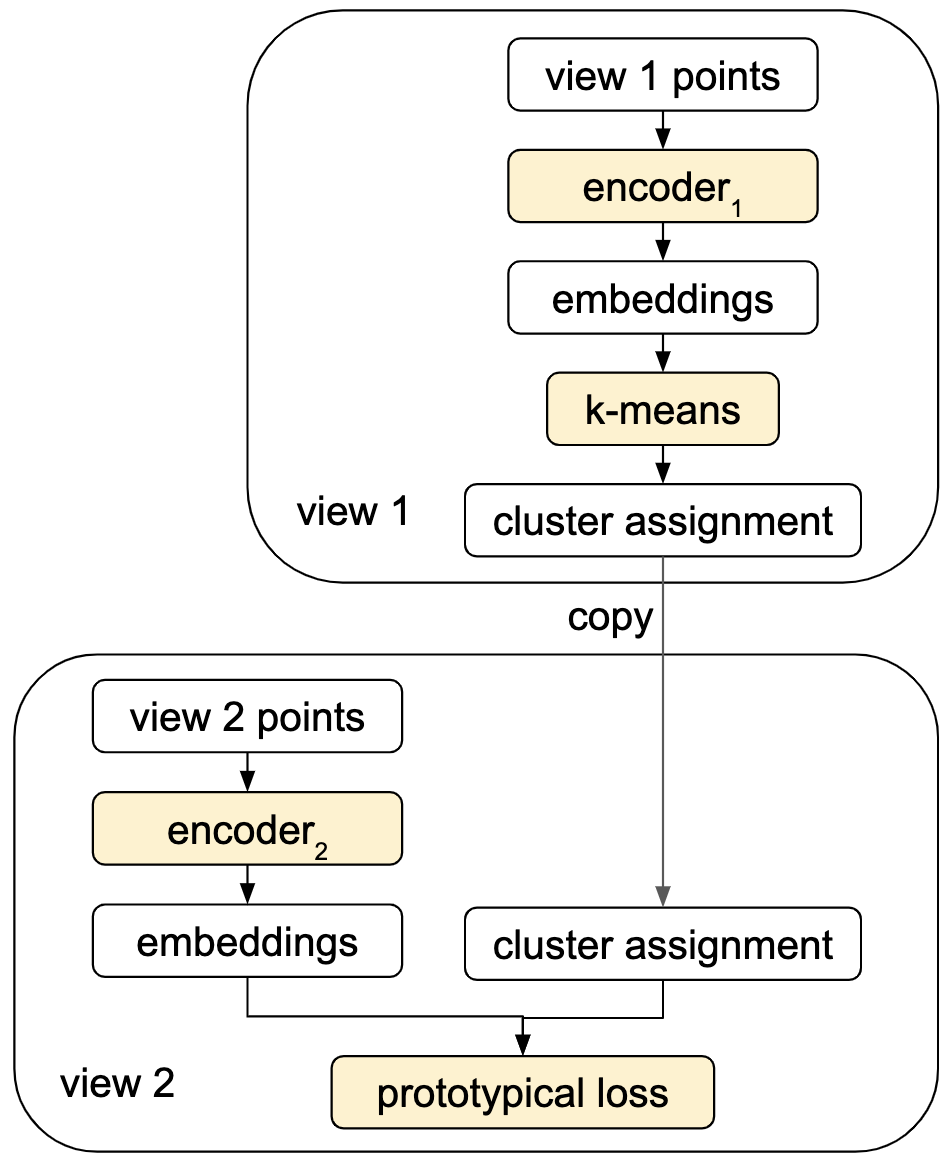}
\caption{A depiction of a semi-iteration of the alternating-view k-means algorithm. k-means clustering and prototypical classification are performed for view 1 and view 2 respectively. The view 1 encoder is frozen and the view 2 encoder is updated in this semi-iteration.}
\label{fig:av-kmeans}
\end{figure}

The full training algorithm is presented in~\autoref{algo:av-kmeans}, where $\textproc{K-means}(\{ \mat{x}_i \}, K, M, \{ z'_i \} ) $ is a function that runs k-means clustering on inputs $\{ \mat{x}_i \}$.  $K$ is the number of clusters. $M$ and $\{ z'_i \}$ are optional arguments that represent the number of k-means iterations and the initial cluster assignment. The function returns cluster assignment $\{ z_i \}$. A visual demonstration of one semi-iteration of~\sysname~is also available in~\autoref{fig:av-kmeans}.

In particular, we initialize the encoders randomly or by using pretrained encoders (\autoref{sec:model:init}). Then, we can obtain the initial cluster assignment by performing k-means clustering on vector representations encoded by $f_{\phi_1}$. During each \sysname~iteration, we first project cluster assignment from view 1 to view 2 and update the neural encoder for view 2 by formulating a supervised learning problem (\autoref{sec:model:train}). Then we perform $M$ vanilla k-means steps to adjust the cluster assignment in view 2 based on the updated encoder. We repeat the procedure for view 2 in the same iteration. Note that in each semi-iteration, the initial centroids corresponding to a view are calculated based on the cluster assignment obtained from the other view. The algorithm runs a total number of $T$ iterations.

\subsection{Prototypical episode training}
\label{sec:model:train}

In each \sysname~iteration, we need to solve two supervised classification problems using the pseudo training datasets $\{(x_i^{(2)},  z_i^{(1)})\}$ and $\{(x_i^{(1)},  z_i^{(2)})\}$ respectively. A simple way to do so is putting a softmax classification layer on top of each encoder network. However, we find that it is beneficial to directly perform classification in the k-means clustering space. To this end, we adopt prototypical networks~\cite{snell2017prototypical}, a metric learning approach, to solely rely on the encoders to form the classifiers instead of introducing additional classification layers.  

Given input data $\{ (x_i, z_i)  \}$ and a neural network encoder $f_{\phi}$, prototypical networks compute a $D$-dimensional representation $\mat{c}_k$, or prototype, of each class by averaging the vectors of the embedded support points belonging to its class:
\begin{equation}
\mat{c}_k = \frac{1}{|S_k|}\sum_{(x_i, z_i) \in S_k} f_{\phi} (x_i)
\end{equation}
here we drop the view superscripts for simplicity. Conceptually, the prototypes $\{ \mat{c}_k \}$ are similar to the centroids in the k-means algorithm, except that a prototype is computed on a subset of the instances of a class (the support set) while a centroid is computed based on all instances of a class.

Given a sampled query data point $x$, prototypical networks produce a distribution over classes based on a softmax over distances to the prototypes in the embedding space:
\begin{equation}
p (y = k | x) = \frac{\exp (- d( f_{\phi}(x),  \mat{c}_k  ) ) }{\sum_{k'} \exp (- d( f_{\phi}(x),  \mat{c}_{k'}  ) ) },
\label{eq:softmax}
\end{equation}
where the distance function is the squared Euclidean distance $d(\mat{x}, \mat{x'}) = ||\mat{x} - \mat{x'}||^2$.

The model minimizes the negative log-likelihood of the data: $L(\phi) = - \log p (y = k | x) $. Training episodes are formed by randomly selecting a subset of classes from the training set, then choosing a subset of examples within each class to act as the support set and a subset of the remainder to serve as query points. We refer to the original paper~\cite{snell2017prototypical} for more detailed description of the model.


\subsection{Parameter initialization}
\label{sec:model:init}

Although \sysname~can effectively work with random parameter initializations, we do expect that it will benefit from initializations obtained from pretrained models with some well-studied unsupervised learning objectives. We present two methods to initialize the utterance encoders for both the query and content views. The first approach is based on recurrent autoencoders. We embed an utterance using a BiLSTM encoder. The utterance embedding is then concatenated with every word vector corresponding to the decoder inputs that are fed into a uni-directional LSTM decoder. We use the neural encoder trained with the autoencoding objective to initialize the two utterance encoders in \sysname.

Recurrent autoencoders independently reconstruct an input utterance without capturing semantic dependencies across consecutive utterances. We consider a second initialization method, quick thoughts~\cite{logeswaran2018efficient}, that addresses the problem by predicting a context utterance from a set of candidates given a target utterance. Here, the target utterances are sampled randomly from the corpus, and the context utterances are sampled from within each pair of adjacent utterances. We use two separate BiLSTM encoders to encode utterances, which are named as the target encoder $f$ and the context encoder $g$. To score the compatibility of a target utterance $\mat{s}$ and a candidate context utterance $\mat{t}$, we simply use the inner product of the two utterance vectors $f(\mat{s})^\top \cdot g(\mat{t})$. The training objective maximizes the log-likelihood of the context utterance given the target utterance and the candidate utterance set. After pretraining, we adopt the target encoder to initialize the two utterance encoders in \sysname.

\section{Data}
\label{sec:data}

As discussed in the introduction, existing goal-oriented dialog datasets mostly concern predefined dialog intents in some narrow domains such as restaurant or travel booking~\cite{henderson2014second, budzianowski2018multiwoz, serban2018survey}. To carry out this study, we adopt a more challenging corpus that consists of human-human conversations for customer service and manually annotate the user intents of a small number of dialogs. We also build a question intent clustering dataset to assess the generalization ability of the proposed method on the related problem.

\subsection{Twitter airline customer support}

\begin{table*}
\small
\centering
    \begin{tabular}{lll}
    \toprule
        Dialog intent & \# Dialogs & Query utterance example \\
         \midrule
         \texttt{Baggage} & 40 & hi, do suit bags count as a personal items besides carry on baggage? \\
         \texttt{BookFlight} & 27 & trying all day to book an international flight, only getting error msg. \\
         \texttt{ChangeFlight} & 16 & can i request to change my flight from lax to msy on 10/15? \\
         \texttt{CheckIn} & 21 & hy how can i have some help... having some problems with a check in \\
         \texttt{CustomerService} & 19 &  2 hour wait time to talk to a customer service agent?!? \\
         \texttt{FlightDelay} & 85 & delay... detroit $<$ orlando \\
         \texttt{FlightEntertainment} & 40 & $@$airline is killing it with these inflight movie options \\
         \texttt{FlightFacility} & 32 & just flew $@$airline economy... best main cabin seat ive ever sat in. \\
         \texttt{FlightStaff} & 30 & great crew on las vegas to baltimore tonight. \\
         \texttt{Other} & 116 & hi, i have a small question! \\
         \texttt{RequestFeature} & 10 & when are you going to update your app for iphone x? \\
         \texttt{Reward} & 17 & need to extend travel funds that expire tomorrow! \\
         \texttt{TerminalFacility} & 13 & thx for the new digital signs at dallas lovefield. well done!! \\
         \texttt{TerminalOperation} & 34 & would be nice if you actually announced delays \\
    \bottomrule
    \end{tabular}
    \caption{Statistics of the labeled Twitter airline customer support (TwACS) dataset and the corresponding user query utterance examples. The Twitter handles of the airlines are replaced by $@$airline.}
    \label{tab:airline} 
\end{table*}

We consider the customer support on Twitter corpus released by Kaggle,\footnote{\url{https://www.kaggle.com/thoughtvector/customer-support-on-twitter}} which contains more than three million tweets and replies in the customer support domain. The tweets constitute conversations between customer support agents of some big companies and their customers. As the conversations regard a variety of dynamic topics, they serve as an ideal testbed for the dialog intent induction task. In the customer service domain, different industries generally address unrelated topics and concerns. We focus on dialogs in the airline industry,\footnote{We combined conversations involved the following Twitter handles: $@$Delta, $@$British\_Airways, $@$SouthwestAir, and $@$AmericanAir.} as they represent the largest number of conversations in the corpus. We name the resulting dataset the \emph{Twitter airline customer support (TwACS)} corpus. We rejected any conversation that redirects the customer to a URL or another communication channel, e.g., direct messages. We ended up with a dataset of $43,072$ dialogs. The total numbers of dialog turns and tokens are $63,147$ and $2,717,295$ respectively.

After investigating $500$ randomly sampled conversations from TwACS, we established an annotation task with $14$ dialog intents and hired two annotators to label the sampled dialogs based on the user query utterances. The Cohen's kappa coefficient was $0.75$, indicating a substantial agreement between the annotators. The disagreed items were resolved by a third annotator. To our knowledge, this is the first dialog intent induction dataset. The data statistics and user query utterance examples corresponding to different dialog intents are presented in~\autoref{tab:airline}.

\subsection{AskUbuntu}

\emph{AskUbuntu} is a dataset collected and processed by~\newcite{shah2018adversarial} for the duplicate question detection task. The dataset consists of technical support questions posted by users on AskUbuntu website with annotations indicating that two questions are semantically equivalent. For instance,

\paragraph{$q_1:$} {\it how to install ubuntu w/o removing windows}
\vspace{-0.3in}
\paragraph{$q_2:$} {\it installing ubuntu over windows 8.1}
\vspace{0.1in}

\noindent are duplicate and they can be resolved with similar answers. A total number of $257,173$ questions are included in the dataset and $27,289$ pairs of questions are labeled as duplicate ones. In addition, we obtain the top rated answer for each question from the AskUbuntu website dump.\footnote{\url{https://archive.org/details/stackexchange}} 

In this work, we reprocess the data and build a question intent clustering dataset using an automatic procedure. Following~\newcite{haponchyk2018supervised}, we transform the duplicate question annotations into the question intent cluster annotations with a simple heuristic: for each question pair $q_1$, $q_2$ annotated as a duplicate, we assigned $q_1$ and $q_2$  to the same cluster. As a result, the question intent clusters correspond to the connected components in the duplicate question graph. There are $7,654$ such connected components. However, most of the clusters are very small: $91.7\%$ of the clusters contain only $2$--$5$ questions. Therefore, we experiment with the largest $20$ clusters. After filtering out questions without corresponding answers, there are $833$ questions in this work. The sizes of the largest and the smallest clusters considered in this study are $284$ and $11$ respectively.

\section{Experiments}
\label{sec:exp}

In this section, we evaluate \sysname~on the TwACS and AskUbuntu datasets as described in~\autoref{sec:data}. We compare \sysname~with competitive systems for representation learning or multi-view clustering and present our main findings in~\autoref{sec:exp:results}. In addition, we examine the output clusters obtained from \sysname~on the TwACS dataset to perform a thoughtful error analysis.

\begin{table*} [ht!]
\centering
\small
\begin{tabular}{llcllllllll}
    \toprule
    \multirow{2}{*}{\parbox{2cm}{Clustering\\algorithm}} & \multirow{2}{*}{\parbox{2cm}{Pretraining\\method}} & \multicolumn{4}{c}{TwACS} & \multicolumn{4}{c}{AskUbuntu} \\
    \cmidrule(l){3-6} \cmidrule(l){7-10}
          & & Prec & Rec & F1 & ACC & Prec & Rec & F1 & ACC \\ \midrule
   \multicolumn{10}{l}{\it Baseline systems} \\
     k-means & PCA &  28.1 & 28.3 & 28.2 & 19.8 & 35.1 & 27.8 & 31.0 & 22.0 \\
     & autoencoders &  34.4 & 25.9 & 29.5 & 23.2 & 27.3 & 20.1 & 23.1 & 14.6 \\
     & quick thoughts & 46.7 & 38.3 & 42.1 & 35.4 & 42.9 & 39.4 & 41.1 & 33.1 \\[3pt]
     MVSC & PCA & 32.4 & 24.2 & 27.8 & 22.6 & 40.4 & 27.7 & 32.9 & 25.5 \\
     & autoencoders & 36.1 & 27.7 & 31.3 & 24.9 & 36.7 & 22.8 & 28.2 & 21.4 \\
     & quick thoughts & 45.8 & 35.4 & 40.0 & 32.6 & 35.1 & 23.4 & 28.1 & 22.3 \\[6pt]
   \multicolumn{10}{l}{\it Our approach}\\
     \sysname~& no pretraining  & 37.5 & 33.6 & 35.4 & 29.5 & 52.0 & 51.9 & 51.9 & \textbf{44.0}  \\
     & autoencoders & 44.4 & 34.6 & 38.9 & 31.6 & 50.6 & 46.1 & 48.2 & 39.7  \\
     & quick thoughts & \textbf{48.9} & \textbf{43.8} & \textbf{46.2} & \textbf{39.9} & \textbf{53.8} & \textbf{52.7} & \textbf{53.3} & 41.1  \\
    \bottomrule
\end{tabular}
\caption{Evaluation results on the TwACS and AskUbuntu datasets for different systems. MVSC is short for the multi-view spectral clustering algorithm proposed by~\newcite{kanaan2018multiview}. The pretrained representations are fixed during k-means and MVSC clustering and they are fine-tuned during \sysname~clustering. The best results are in {\bf bold}.}
\label{tab:results}
\end{table*}

\subsection{Experimental settings}

We train the models on all the instances of a dataset and evaluate on the labeled instances. We employ the publicly available 300-dimensional GloVe vectors~\cite{pennington2014glove} pretrained with 840 billion tokens to initialize the word embeddings for all the models.

\paragraph{Competitive systems}
We consider state-of-the-art methods for representation learning and/or multi-view clustering as our baseline systems. We formulate the dialog induction task as an unsupervised clustering task and include two popular clustering algorithms \emph{k-means} and \emph{spectral clustering}. \emph{multi-view spectral clustering (MVSC)}~\cite{kanaan2018multiview} is a competitive standard multi-view clustering approach.\footnote{We use the scikit-learn k-means implementation and the MVSC implementation available at: \url{https://pypi.org/project/multiview/}.} In particular, we carry out clustering using the query-view and content-view representations learned by the representation learning methods (k-means only requires query-view representations). In the case where a content-view input corresponds to multiple utterances, we take the average of the utterance vectors as the content-view output representation for autoencoders and quick thoughts. 

\sysname~is a joint representation learning and multiview clustering method. Therefore, we compare with SOTA representation learning methods \emph{autoencoders}, and \emph{quick thoughts}~\cite{logeswaran2018efficient}. Quick thoughts is a strong representation learning baseline that is adopted in BERT~\cite{bert}. We also include \emph{principal component analysis (PCA)}, a classic representation learning and dimensionality reduction method, since bag-of-words representations are too expensive to work with for clustering analysis.

We compare three variants of \sysname~that differ in the pretraining strategies. In addition to the \sysname~systems pretrained with autoencoders and quick thoughts, we also consider a system whose encoder parameters are randomly initialized (\emph{no pretraining}). 

\paragraph{Metrics}
We compare the competitive approaches on a number of standard evaluation measures for clustering analysis. Following prior work~\cite{kumar2011co, haponchyk2018supervised, xie2016unsupervised}, we set the number of clusters to the number of ground truth categories and report precision, recall, F1 score, and unsupervised clustering accuracy (ACC). To compute precision or recall, we assign each predicted cluster to the most frequent gold cluster or assign each gold cluster to the most frequent predicted cluster respectively. The F1 score is the harmonic average of the precision and recall. ACC uses a one-to-one assignment between the gold standard clusters and the predicted clusters. The assignment can be efficiently computed by the Hungarian algorithm~\cite{kuhn1955hungarian}.

\paragraph{Parameter tuning}
We empirically set both the dimension of the LSTM hidden state and the number of principal components in PCA to $300$. The number of \sysname~iterations $T$ and the number of k-means steps in a \sysname~semi-iteration $M$ are set to $50$ and $10$ respectively, as we find that more iterations lead to similar cluster assignments. We adopt the same set of hyperparameter values as used by~\newcite{snell2017prototypical} for training the prototypical networks. Specifically, we fix the number of query examples and the number of support examples to $15$ and $5$. The networks are trained for $100$ episodes per \sysname~semi-iteration. The number of sampled classes per episode is chosen to be $10$, as it has to be smaller than the number of ground truth clusters. Adam~\cite{kingma2015adam} is utilized to optimize the models and the initial learning rate is $0.001$.  During autoencoders or quick thoughts pretraining, we check the performance on the development set after each epoch to perform early stopping, where we randomly sample 10\% unlabeled instances as the development data.

\subsection{Results}
\label{sec:exp:results}
Our main empirical findings are presented in~\autoref{tab:results}, in which we compare \sysname~with standard single-view and multi-view clustering algorithms. We also evaluate classic and neural approaches for representation learning, where the pretrained representations are fixed during k-means and MVSC clustering and they are fine-tuned during \sysname~clustering. We analyze the empirical results in details in the following paragraphs.

\paragraph{Utilizing multi-view information} 
Among all the systems, k-means clustering on representations trained with PCA or autoencoders only employs single-view information encoded in user query utterances. They clearly underperform the rest of the systems that leverage multi-view information of the entire conversations.  Quick thoughts infuses the multi-view knowledge through the learning of the query-view vectors that are aware of the content-view semantics. In contrast, multi-view spectral clustering can work with representations that are separately learned for the individual views and the multi-view information is aggregated using the common eigenvectors of the data similarity Laplacian matrices. As shown, k-means clustering on quick thoughts vectors gives superior results than MVSC pretrained with PCA or autoencoders by more than $10\%$ F1 or ACC, which indicates that multi-view representation learning is effective for problems beyond simple supervised learning tasks. Combining representation learning and multi-view clustering in a static way seems to be less ideal---MVSC performs worse than k-means using the quick thoughts vectors as clustering inputs. Multi-view representation learning breaks the independent-view assumption that is critical for classic multi-view clustering algorithms.

\paragraph{Joint representation learning and clustering}
We now investigate whether joint representation learning and clustering can reconcile the conflict between cross-view representation learning and classic multi-view clustering. \sysname~outperforms k-means and MVSC baselines by considerable margins. It achieves $46\%$ and $53\%$ F1 scores and $40\%$ and $44\%$ ACC scores on the TwACS and AskUbuntu datasets, which are $5$--$30$ percent higher than competitive systems. Compared to alternative methods, \sysname~is able to effectively seek clustering-friendly representations that also encourage similar cluster assignments for different views of the same instances. With the help of quick thoughts pretraining, \sysname~improves upon the strongest baseline, k-means clustering on quick thoughts vectors, by $4.5\%$ ACC on the TwACS dataset and $12.2\%$ F1 on the AskUbuntu dataset.

\paragraph{Model pretraining for \sysname}
Evaluation results on \sysname~with different parameter initialization strategies are available in~\autoref{tab:results}. As suggested, pretraining neural encoders is important for obtaining competitive results on the TwACS dataset, while its impact on the AskUbuntu dataset is less pronounced. AskUbuntu is six times larger than TwACS and models trained on AskUbuntu are less sensitive to their parameter initializations. This observation is consistent with early research on unsupervised pretraining, where~\newcite{schmidhuber2012multi} argue that unsupervised initialization/pretraining is not necessary if a large amount of training data is available. Between the two pretraining methods, quick thoughts is much more effective than autoencoders---it improves upon no pretraining and autoencoders by $10.4\%$  and $8.3\%$ ACC scores on the TwACS dataset.

\subsection{Error analysis}

\begin{table} [ht!]
\centering
\small
\addtolength{\tabcolsep}{-2pt}
\begin{tabular}{lll}
    \toprule
    Ground truth & Prediction & \# Instances \\ \midrule
     \texttt{Other} & \texttt{CustomerService} & 21 \\
     \texttt{TerminalOp.} & \texttt{FlightDelay} & 10 \\
     \texttt{FlightDelay} & \texttt{ChangeFlight} & 10 \\
     \texttt{Other} & \texttt{FlightStaff} & 10 \\
     \texttt{FlightEnter.} & \texttt{Other} & 8 \\
    \bottomrule
\end{tabular}
\caption{The top 5 most frequent errors made by the quick thoughts pretrained \sysname~on the TwACS dataset. The one-to-one assignment between the gold clusters and the predicted clusters is computed by the Hungarian algorithm.}
\label{tab:error}
\end{table}

Our best performed system still fails to hit $50\%$ F1 or ACC score on the TwACS dataset. We examine the outputs of the quick thoughts pretrained \sysname~on TwACS, focusing on investigating the most frequent errors made by the system. To this end, we compute the confusion matrix based on the one-to-one assignment between the gold clusters and the predicted clusters used by ACC. The top 5 most frequent errors are presented in~\autoref{tab:error}. As shown, three of the five errors involve \texttt{Other}.  Instances under the \texttt{Other} category correspond to miscellaneous dialog intents, thereby they are less likely to be grouped together based on the semantic meaning representations.

The other two frequent errors confuse \texttt{FlightDelay} with \texttt{TerminalOperation} and \texttt{ChangeFlight} respectively. Poor terminal operations often incur unexpected customer delays. Two example query utterances are shown as follows,

\paragraph{$q_1:$} {\it who’s running operation at mia flight 1088 been waiting for a gate.}
\vspace{-0.1in}
\paragraph{$q_2:$} {\it have been sitting in the plane waiting for our gate for 25 minutes.}
\vspace{0.1in}

\noindent Sometimes, a user may express more than one intents in a single query utterance. For example, in the following query utterance, the user complaints about the delay and requests for an alternative flight:

\paragraph{$q:$} {\it why is ba flight 82 from abuja to london delayed almost 24 hours? and are you offering any alternatives?}
\vspace{0.1in}

\noindent We leave multi-intent induction to future work.

\section{Related Work}
\label{sec:related}

\paragraph{User intent clustering}

Automatic discovery of user intents by clustering user utterances is a critical task in understanding the dynamics of a domain with user generated content. Previous work focuses on grouping similar web queries or user questions together using supervised or unsupervised clustering techniques. \newcite{kathuria2010classifying} perform simple k-means clustering on a variety of query traits to understand user intents. \newcite{cheung2012sequence} present an unsupervised method for query intent clustering that produces a pattern consisting of a sequence of semantic concepts and/or lexical items for each intent. \newcite{jeon2005finding} use machine translation to estimate word translation probabilities and retrieve similar questions from question archives. A variation of k-means algorithm, MiXKmeans, is presented by~\newcite{deepak2016mixkmeans} to cluster threads that present on forums and Community Question Answering websites. \newcite{haponchyk2018supervised} propose to cluster questions into intents using a supervised learning method  that yields better semantic similarity modeling. Our work focuses on a related but different task that automatically induces user intents for building dialog systems. Two sources of information are naturally available for exploring our deep multi-view clustering approach.

\paragraph{Multi-view clustering}
Multi-view clustering (MVC) aims at grouping similar subjects into the same cluster by combining the available multi-view feature information to search for consistent cluster assignments across different views~\cite{chao2017survey}. 
Generative MVC approaches assume that the data is drawn from a mixture model and the membership information can be inferred using the multi-view EM algorithm~\cite{bickel2004multi}. Most of the works on MVC employ discriminative approaches that directly optimize an objective function that involves pairwise similarities so that the average similarity within clusters can be minimized and the average similarity between clusters can be maximized. In particular, \newcite{chaudhuri2009multi} propose to exploit canonical correlation analysis to learn multi-view representations that are then used for downstream clustering. Multi-view spectral clustering~\cite{kumar2011co, kanaan2018multiview} constructs a similarity matrix for each view and then iteratively updates a matrix using the eigenvectors of the similarity matrix computed on another view. Standard MVC algorithms expect multi-view feature inputs that are fixed during unsupervised clustering. \sysname~works with raw multi-view text inputs and learns representations that are particularly suitable for clustering.

\paragraph{Joint representation learning and clustering}
Several recent works propose to jointly learn feature representations and clustering via neural networks. \newcite{xie2016unsupervised} present the deep embedded clustering (DEC) method that learns a mapping from the data space to a lower-dimensional
feature space where it iteratively optimizes a KL divergence based clustering objective. Deep clustering network (DCN)~\cite{yang2017towards} is a joint dimensional reduction and k-means clustering framework, in which the dimensional reduction model is implemented with a deep neural network. These methods focus on the learning of single-view representations and the multi-view information is under-explored. \newcite{lin2018jointly} present a joint framework for deep multi-view clustering (DMJC) that is the closest work to ours. However, DMJC only works with single-view inputs and the feature representations are learned using a multi-view fusion mechanism. In contrast, \sysname~assumes that the inputs can be naturally partitioned into multiple views and carry out learning with the multi-view inputs directly.




\section{Conclusion}
\label{sec:con}

We introduce the novel task of dialog intent induction that concerns automatic discovery of dialog intents from user query utterances in human-human conversations. The resulting dialog intents provide valuable insights in helping design goal-oriented dialog systems. We propose to leverage the dialog structure to divide a dialog into two independent views and then present \sysname, a deep multi-view clustering algorithm, to jointly perform multi-view representation learning and clustering on the views. We conduct extensive experiments on a Twitter conversation dataset and a question intent clustering dataset. The results demonstrate the superiority of \sysname~over competitive representation learning and multi-view clustering baselines. In the future, we would like to abstract multi-view data from multi-lingual and multi-modal sources and investigate the effectiveness of \sysname~on a wider range of tasks in the multi-lingual or multi-modal settings.

\section*{Acknowledgments}

We thank Shawn Henry and Ethan Elenberg for their comments on an early draft of the paper. We also thank the NLP team and the General Research team of ASAPP for their support throughout the project. We thank the EMNLP reviewers for their
helpful feedback. We thank Willie Chang for his post-publication bug reports.

\bibliographystyle{acl_natbib}
\bibliography{cite-strings,cites,cite-definitions}

\clearpage
\appendix
\onecolumn
\section{Appendix: Results of Full Datasets}
\label{sec:appendix}

In the original paper, we stated that there were 4,692 duplicate questions in the askubuntu dataset, after selecting the top 20 clusters. However, we further filtered these questions according to whether the answer was available, giving 1,029 question-answer pairs. We accidentally partitioned this further, and used 833 question-answer pairs as the labeled evaluation dataset. In the experiments below, we use the full 1,029 QA pairs evaluation dataset. We additionally use the latest version of the code, after fixing some issues that we noticed post-publication.

\begin{table*} [hb!]
\centering
\small
\begin{tabular}{llcllllllll}
    \toprule
    \multirow{2}{*}{\parbox{2cm}{Clustering\\algorithm}} & \multirow{2}{*}{\parbox{2cm}{Pretraining\\method}} & \multicolumn{4}{c}{TwACS} & \multicolumn{4}{c}{AskUbuntu} \\
    \cmidrule(l){3-6} \cmidrule(l){7-10}
          & & Prec & Rec & F1 & ACC & Prec & Rec & F1 & ACC \\ \midrule
\multicolumn{10}{l}{\it Baseline systems} \\
k-means & PCA & 27.1 & 26.7 & 26.9 & 18.3 & 33.0 & 32.9 & 32.9 & 20.2 \\
 & autoencoders & 27.1 & 18.7 & 22.2 & 15.9 & 27.9 & 19.3 & 22.8 & 14.4 \\
 & quick thoughts & 48.1 & 37.5 & 42.1 & 34.6 & 45.5 & 44.6 & 45.0 & 34.4 \\
[3pt]
MVSC & PCA & 31.6 & 24.2 & 27.4 & 22.6 & \textbf{70.8} & 41.7 & 52.5 & 37.5 \\
 & autoencoders & 29.3 & 21.2 & 24.6 & 18.9 & 28.6 & 13.0 & 17.9 & 11.2 \\
 & quick thoughts & 45.6 & 34.8 & 39.5 & 33.2 & 53.4 & 37.3 & 43.9 & 34.5 \\
[6pt] \multicolumn{10}{l}{\it Our approach}\\
AV-KMeans & no pretraining & 38.3 & 27.3 & 31.9 & 26.5 & 52.2 & 50.3 & 51.3 & 40.7 \\
 & autoencoders & 46.0 & 33.6 & 38.9 & 31.6 & 50.2 & 43.8 & 46.8 & 36.2 \\
 & quick thoughts & \textbf{51.1} & \textbf{38.1} & \textbf{43.7} & \textbf{35.9} & 56.6 & \textbf{55.1} & \textbf{55.8} & \textbf{43.2} \\
 \bottomrule
\end{tabular}
\caption{Evaluation results on the TwACS and AskUbuntu datasets for different systems. MVSC is short for the multi-view spectral clustering algorithm proposed by~\newcite{kanaan2018multiview}. The pretrained representations are fixed during k-means and MVSC clustering and they are fine-tuned during \sysname~clustering. The best results are in {\bf bold}.}
\label{tab:apx_results}
\end{table*}

\end{document}